\documentclass[letterpaper]{article} 
\usepackage{aaai24_preprint}  
\usepackage{times}  
\usepackage{helvet}  
\usepackage{courier}  
\usepackage[hyphens]{url}  
\usepackage{graphicx} 
\usepackage{natbib}  
\usepackage{caption} 
\usepackage{algorithm}
\usepackage{algorithmic}

\usepackage{multirow}
\usepackage{booktabs}
\usepackage{amssymb}
\usepackage{amsmath}
\usepackage{color}

\usepackage[capitalize]{cleveref}
\crefname{section}{Sec.}{Secs.}
\Crefname{section}{Section}{Sections}
\Crefname{table}{Table}{Tables}
\crefname{table}{Tab.}{Tabs.}
\usepackage{newfloat}
\usepackage{listings}
\DeclareCaptionStyle{ruled}{labelfont=normalfont,labelsep=colon,strut=off} 
\floatstyle{ruled}
\newfloat{listing}{tb}{lst}{}
\floatname{listing}{Listing}
\title{Layout and Task Aware Instruction Prompt for Zero-shot Document Image Question Answering}
\author{
    Wenjin Wang, Yunhao Li, Yixin Ou, Yin Zhang\thanks{Corresponding author: Yin Zhang.}
}
\affiliations{
    Zhejiang University, Hangzhou, China\\

    \{wangwenjin,12121078,ouyixin,zhangyin98\}@zju.edu.cn
}

\usepackage{bibentry}

\begin{document}

\maketitle

\begin{abstract}
Layout-aware pre-trained models has achieved significant progress on document image question answering.
They introduce extra learnable modules into existing language models to capture layout information within document images from text bounding box coordinates obtained by OCR tools.
However, extra modules necessitate pre-training on extensive document images.
This prevents these methods from directly utilizing off-the-shelf instruction-tuning language foundation models, which have recently shown promising potential in zero-shot learning.
Instead, in this paper, we find that \emph{instruction-tuning language models like Claude and ChatGPT can understand layout by spaces and line breaks}.
Based on this observation, we propose the \textbf{LA}yout and \textbf{T}ask aware \textbf{In}struction \textbf{Prompt} (\textbf{LATIN-Prompt}), which consists of layout-aware document content and task-aware instruction.
Specifically, the former uses appropriate spaces and line breaks to recover the layout information among text segments obtained by OCR tools, and the latter ensures that generated answers adhere to formatting requirements.
Moreover, we propose the \textbf{LA}yout and \textbf{T}ask aware \textbf{In}struction \textbf{Tuning} (\textbf{LATIN-Tuning}) to improve the performance of small instruction-tuning models like Alpaca.
Experimental results show that LATIN-Prompt enables zero-shot performance of Claude and ChatGPT to be comparable to the fine-tuning performance of SOTAs on document image question answering, and LATIN-Tuning enhances the zero-shot performance of Alpaca significantly.
For example, LATIN-Prompt improves the performance of Claude and ChatGPT on DocVQA by $263\%$ and $20\%$ respectively.
LATIN-Tuning improves the performance of Alpaca on DocVQA by $87.7\%$.
Quantitative and qualitative analyses demonstrate the effectiveness of LATIN-Prompt and LATIN-Tuning.
We release our code to facilitate future research\footnote{https://github.com/WenjinW/LATIN-Prompt}.
\end{abstract}

\section{Introduction}
\begin{figure}
\centering
\small
\includegraphics[width=1\linewidth]{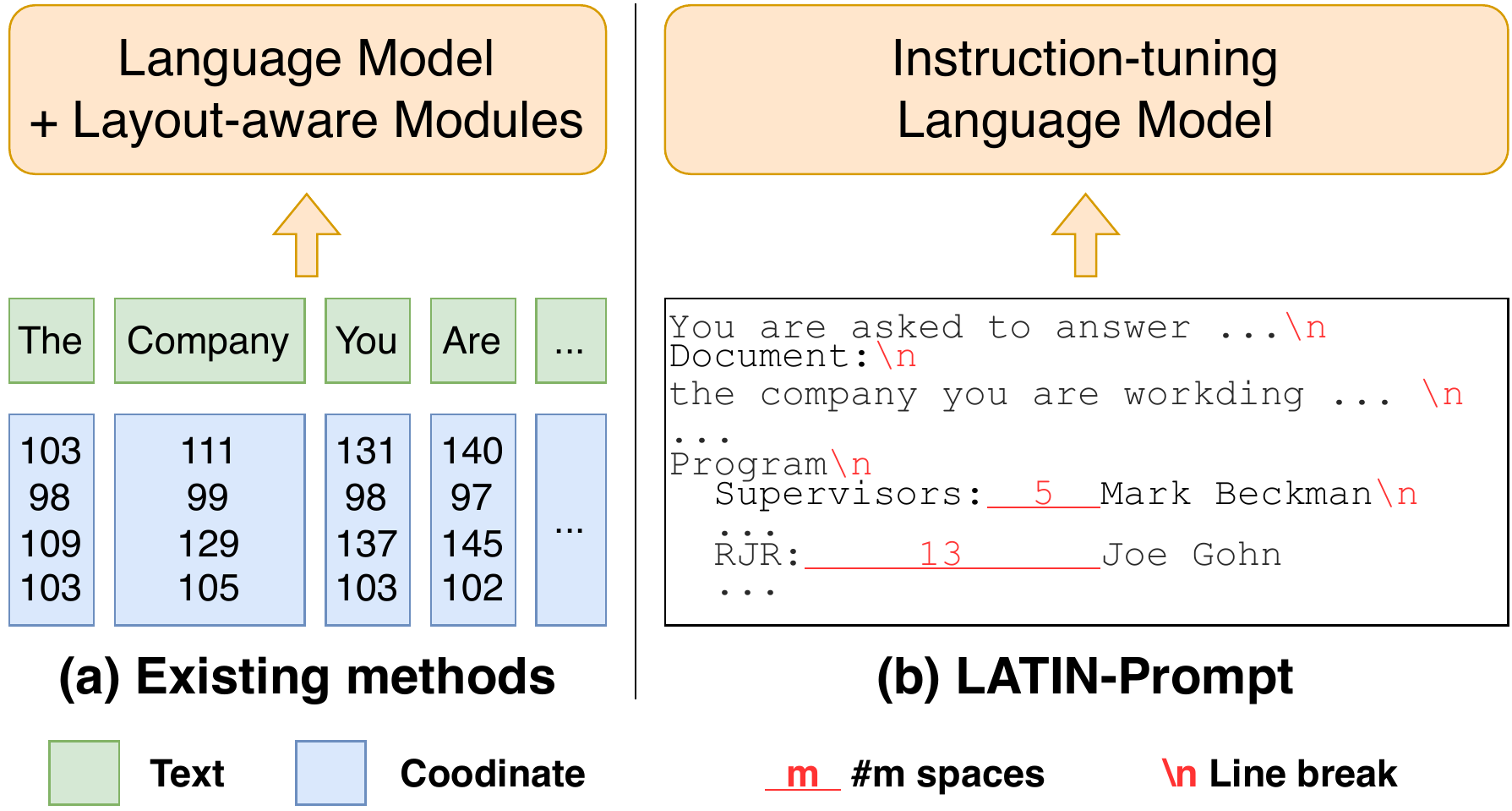}
\caption{
(a) Existing methods introduce layout-aware modules into language models to capture layout information within document images from text bounding box coordinates obtained by OCR tools.
They need further pre-training on extensive document images.
(b) Our method allows instruction-tuning language models to \emph{capture layout by spaces and line breaks} and can conduct \emph{zero-shot inference} on document image question-answering.
}
\label{fig:motivation}
\end{figure}
Intelligent document image question answering, as an important application of document intelligence, aims to develop AI systems to automatically answer natural language questions based on the understanding of document images.
Compared with text documents, document images contain textual, visual, and layout information, which pose unique challenges for machine comprehension.

Recently, layout-aware pre-trained models has achieved significant progress on document image question answering.
They introduce extra learnable modules on top of language models~\cite{devlinBERTPretrainingDeep2019,liuRoBERTa2019,raffelT52020,baoUniLMv22020} to capture layout information within document images from coordinates of text bounding box obtained by OCR tools (\cref{fig:motivation}(a)).
LayoutLM~\cite{xuLayoutLM2020} introduces coordinate information into model input by 2D position embeddings. LayoutLMv2~\cite{xuLayoutLMv22021}, LayoutLMv3~\cite{huangLayoutLMv32022}, and ERNIE-Layout~\cite{pengERNIELayout2022} capture the layout information from coordinate by layout-aware attention mechanism.
These methods conduct pre-training on extensive document images for the newly introduced layout-aware modules.

However, the need of pre-training prevents these methods from directly utilizing off-the-shelf instruction-tuning language foundation models, which have recently shown promising potential in zero-shot learning.
On the one hand, commercial large instruction-tuning models like Claude~\cite{Claude} and ChatGPT~\cite{ChatGPT} are closed-source, impeding further pre-training.
On the other hand, open-source instruction-tuning models are of much larger scale than traditional models.
For example, Alpaca~\cite{StanfordAlpacaInstructionfollowing2023} consists of 7 billion parameters, whereas BERT$_\text{large}$~\cite{devlinBERTPretrainingDeep2019} only comprises 300+ million parameters.
Existing methods~\cite{xuLayoutLM2020,xuLayoutLMv22021,huangLayoutLMv32022,pengERNIELayout2022} select over 10 million pages from the IIT-CDIP Test Collection dataset~\cite{lewisIIT-CDIP2006} for pre-training. But the cost of pre-training instruction-tuning models like Alpaca on 10 million pages is too expensive.

Instead, in this work, we find that \emph{instruction-tuning language models like Claude and ChatGPT can understand layout by spaces and line breaks} (\cref{fig:motivation}(b)).
Based on this observation, we propose the \textbf{LA}yout and \textbf{T}ask aware \textbf{In}struction \textbf{Prompt} (\textbf{LATIN-Prompt}), which consists of layout-aware document content and task-aware instruction.
Specifically, given the OCR results, we use appropriate spaces and line breaks to connect all the text segments together, resulting in the layout-aware document content.
The layout information contained within the coordinates is translated into spaces and line breaks.
Further, we integrate task instruction into layout-aware document content, ensuring that the model generates answers that adhere to the formatting requirements.
Although simple, our method is intuitive and consistent with human behavior.
Humans employ whitespace (blank) regions to represent and comprehend layout.

We also find that small instruction-tuning language foundation models like Alpaca are not good at understanding layout by spaces.
So we propose the \textbf{LA}yout and \textbf{T}ask aware \textbf{In}struction \textbf{Tuning} (\textbf{LATIN-Tuning}) to improve the performance of them.
We convert CSV-format tables into strings containing spaces and line breaks and construct instruction-tuning data from these strings by Claude.

Our contributions are summarized as follows:
\begin{itemize}
\item We find that instruction-tuning models like Claude and ChatGPT can capture layout by spaces and line breaks, and propose LATIN-Prompt to conduct zero-shot inference on document image question-answering tasks.
\item We propose the LATIN-Tuning to enhance the ability of Alpaca to comprehend layout by spaces and line breaks.
\item Experiment results on three datasets show that LATIN-Prompt enables zero-shot performance of Claude and ChatGPT to be comparable to the fine-tuning performance of SOTAs on document image question answering, and LATIN-Tuning enhances the zero-shot performance of Alpaca significantly.
Quantitative and qualitative analyses demonstrate the effectiveness of LATIN-Prompt and LATIN-Tuning.
\end{itemize}

\section{Related Work}
\begin{figure*}[t]
\small
\centering
\includegraphics[width=1\linewidth]{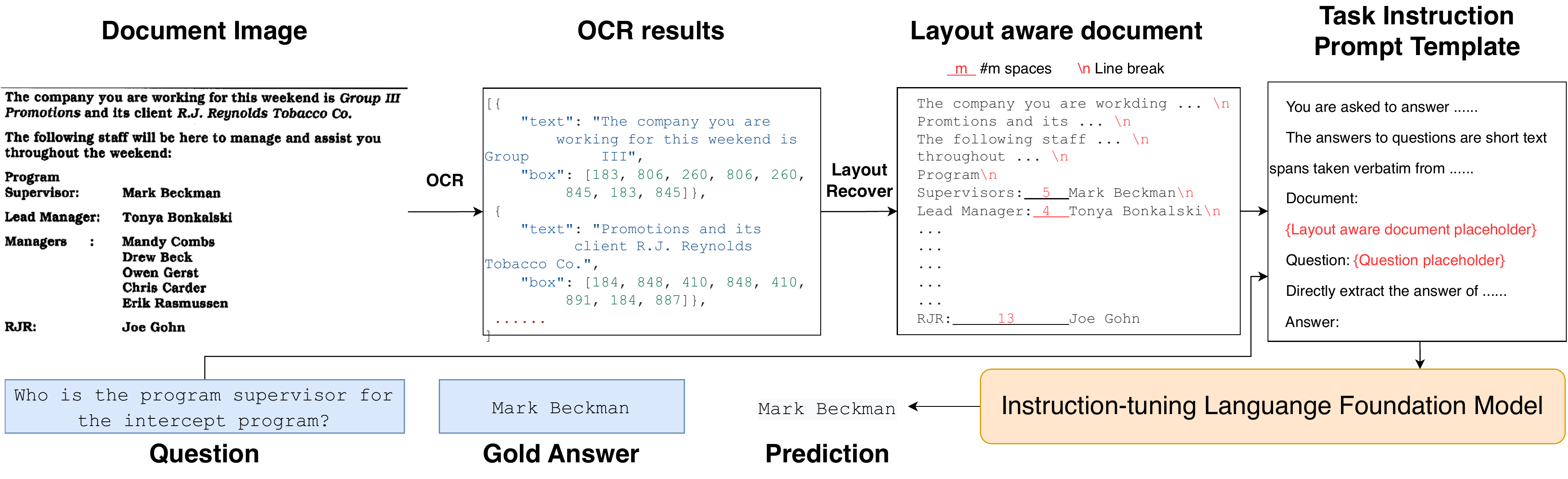}
\caption{
The overview of LATIN-Prompt (\cref{sec:LATIN}).
Given a document image and the corresponding question, we recover the layout information within the document image from OCR results using appropriate spaces and line breaks, and then insert the layout aware document content and question into the task instruction prompt template together.
The instruction-tuning large language foundation model takes the filled template as input and predicts the answer to the question in the required format.
}
\label{fig:LATIN}
\end{figure*}

\begin{table*}[t]
    \small
    \centering
    \begin{tabular}{cl}
    \toprule
    \#Line& Prompt\\
    \midrule
    1& You are asked to answer questions asked on a document image.\\
    2& \parbox[c]{16cm}{
    The answers to questions are short text spans taken verbatim from the document.
    This means that the answers comprise a set of contiguous text tokens present in the document.}\\
    3& Document:\\
    4& \textcolor{red}{\{Layout Aware Document placeholder\}}\\
    5& \\
    6& Question: \textcolor{red}{\{Question placeholder\}}\\
    7&\\
    8&Directly extract the answer of the question from the document with as few words as possible.\\
    9&\\
    10&Answer:\\
    \bottomrule
\end{tabular}
\caption{DocVQA Prompt Template. \textcolor{red}{The \{\} represents the placeholder.}}
\label{tab:docvqa_prompt}
\end{table*}

\subsection{Visually-rich document understanding}
Visually-rich Document Understanding (VrDU) focus on recognizing and understanding scanned or digital-born document images with language, vision, and layout information.
Traditional works in the VrDU employ CNN~\cite{yangLearningExtractSemantic2017,kattiChargridUnderstanding2D2018,denkBERTgridContextualizedEmbedding2019,zhaoCUTIELearningUnderstand2019,sarkhelDeterministicRoutingLayout2019,zhangTRIEEndtoEndText2020,wangRobustVisualInformation2021,linViBERTgridJointlyTrained2021}, GNN~\cite{liuGraphConvolutionMultimodal2019,qianGraphIEGraphBasedFramework2019,yuPICKProcessingKey2020,weiRobustLayoutawareIE2020,carbonellNamedEntityRecognition2021}, and language transformer~\cite{majumderRepresentationLearningInformation2020,wangDocStruct2020} to mine information from document images.

Recently, layout-aware pre-trained Transformers have been proposed \cite{appalarajuDocFormerEndtoEndTransformer2021a,garncarekLAMBERTLayoutAwareLanguage2021a,hwangSpatialDependencyParsing2021,liStructuralLMStructuralPretraining2021,liSelfDoc2021,liStrucTexT2021,xuLayoutXLM2021,hongBROSPreTrainedLanguage2022,leeFormNetStructuralEncoding2022,pengERNIELayout2022,baiWukongReader2022,leePix2Struct2022,luoGeoLayoutLM2023,dhouibDocParserEndtoendOCRfree2023}.
LayoutLM~\cite{xuLayoutLM2020} introduces the 2D position information into the input embedding and LayoutLMv2~\cite{xuLayoutLMv22021} proposes the spatial-aware self-attention mechanism.
Then, LayoutLMv3~\cite{huangLayoutLMv32022} removes the CNN by learning visual features extraction with the discrete image tokens reconstruction, and ERNIE-Layout~\cite{pengERNIELayout2022} introduces the layout knowledge into pre-training.
Further, \cite{zhangSERA2021,guXYLayoutLM2022,wangmmLayout2022} introduce additional designs during the fine-tuning, enabling layout-aware pretrained models to better adapt to downstream tasks.

However, all of existing methods try to understand layout within document images by coordinates of bounding box obtained by OCR tools.
Instead, in this work, we try to directly understand layout by spaces and line breaks.
We propose the LATIN-Prompt and LATIN-Tuning to explore zero-shot document image question answering.

A concurrent work ICL-D3IE~\cite{heICLD3IEInContextLearning2023} also leverages large instruction-tuning models in document image understanding, but it differs significantly from our method.
It focuses on few-shot document information extraction, but in this paper, we focus on zero-shot document image question-answering.

\subsection{Instruction-tuning Language Model}
An ideal AI system should be able to learn and accomplish a variety of tasks according to human instructions.
To this end, many instruction-tuning datasets and language foundation models have been proposed~\cite{thoppilanLaMDALanguageModels2022,chungScalingInstructionFinetunedLanguage2022,weiFLAN2022,sanhT02022,mishraNATURALINSTRUCTIONS2022,wangSUP-NATINST2022,xuMultiInstructImprovingMultiModal2022,iyerOPTIMLScalingLanguage2023,longpreFlanCollectionDesigning2023}.
To fully align models with human intentions and values, reinforcement learning from human feedback and AI feedback~\cite{baiTrainingHelpfulHarmlessAssistantRLHF2022,ouyangInstructGPT2022,baiConstitutionalAIHarmlessness2022,OpenAIGPT42023} are introduced into instruction-tuning.
To reduce the cost of manual annotation of instruction-tuning data, some methods automatically construct instruction-tuning data using off-the-shelf language models~\cite{honovichUnnaturalInstructionsTuning2022,wangSelfInstructAligningLanguage2022,StanfordAlpacaInstructionfollowing2023,pengInstructionTuningGPT42023,zhouAutomaticPromptEngineer(APE)2023,xuWizardLM2023}.
Recently, some methods~\cite{zhuMiniGPT42023,daiInstructBLIP2023} have extended instruction fine-tuning into multimodal domains.

In this paper, we propose the LATIN-Prompt to allow off-the-shelf instruction-tuning language models to conduct zero-shot document image question answering.
Moreover, we propose the LATIN-Tuning to enhance the ability of Alpaca to comprehend layout by spaces and line breaks.

\section{Method}
\label{sec:method}
We propose the LATIN-Prompt and LATIN-Tuning for instruction-tuning language models to conduct zero-shot inference on document image question-answering tasks.

\subsection{LATIN-Prompt}
\label{sec:LATIN}
The key ideas of LATIN-Prompt are as follows:
(1) Capture the layout information by spaces and line breaks
(2) Generate answers adhere to formatting requirements by task instruction.
\cref{fig:LATIN} illustrates the process of LATIN-Prompt.
Given OCR results of a document image, we recover layout information within it by using appropriate spaces and line breaks to connect all the text segments together, resulting in the layout-aware document content.
Then we insert the layout-aware document content and question into the task instruction prompt template.
The instruction-tuning language model takes the filled template as input and predicts the answer to the question in the required format.

Formally, given a document image $D$ and a question answer pair $q$ and $a$, we process the document image by an OCR tool.
The extracted text segments and corresponding bounding boxes are denoted as $S=\{s_1,s_2,\dots,s_n\}$ and $B=\{b_1,b_2,\dots,b_n\}$, where $n$ represents the number of text segments.

\paragraph{Layout Aware Document}
We employ appropriate spaces and line breaks to connect all text segments together, resulting in layout-aware document content.
The process is as follows:

Step 1.
Re-arrange the text segments and bounding boxes in the order from top to bottom and from left to right based on the coordinates.

Step 2.
According to the coordinates, place the text segments and bounding boxes in the $i$-th row into the list $S_i$ and $B_i$ respectively from left to right, and calculate the character count $c_i$ and the width $w_i$ of $i$-th row.
The $w_i$ equals the width of the union of bounding boxes in the list $B_i$.

Step 3.
Calculate the character width of document $D$, which is defined as follows:
\begin{equation}
\label{eq:mean_c}
\bar{c}:=w_i^* / c_i^*, i^*=\text{argmax}_i c_i,
\end{equation}
where $i^*$-row has the maximum character count among all rows.

Step 4.
Join text segments in the same row from left to right by spaces.
Given two adjacent text segments $S_{i,j}$ and $S_{i,k}$, the number of spaces joining them is equal to $h_{i,jk}/\bar{c}$ where $h_{i,jk}$ is the horizontal distance between the two bounding boxes $B_{i,j}$ and $B_{i,k}$.

Step 5.
Join different rows by line breaks to obtain the layout-aware document content (denoted as $S'$).

Recovering layout information by spaces and line breaks is simple but intuitive.
In fact, people do represent and understand layout through blank areas between text elements, rather than precise bounding box coordinates.


\paragraph{Task Aware Instruction}
\label{sec:task}

Different from the open-ended question-answering, document image question-answering typically involves explicit requirements for the answer format.
For example, DocVQA~\cite{mathewDocVQA2021} is an extractive QA task that requires answers to be extracted from the document.
However, with only the layout-aware document content and question, the model can easily generate answers that are not in the document and generate unnecessary descriptions or explanations.

So we integrate task instruction into layout-aware document content, ensuring that the model generates answers that adhere to the formatting requirements.
Specifically, we manually designed different instruction prompt templates for different tasks.
Each template $P(S',q)$ contains the requirement of the task as well as placeholders for the layout-aware document content $S'$ and question $q$.

\Cref{tab:docvqa_prompt} shows the prompt template for DocVQA.
In the first and second lines, we explain the meaning of extraction in detail to the model according to the task description in DocVQA.
Lines 3 to 6 provide placeholders for the layout-aware document and question.
To avoid the model forgetting its task due to the interference of document content, the 8th line summarizes and reiterates the task requirements.
Please refer to the supplementary materials for the prompt templates of InfographicVQA~\cite{mathewInfographicVQA2021} and MP-DocVQA~\cite{titoMulti-PageDocVQAHi-VT52023}.

\paragraph{Zero-shot Inference}
At last, the instruction-tuning language model $M$ takes the filled template $P(S',q)$ as input and predicts the answer as follows:
\begin{equation}
a'=f_{M}(P(S',q)),
\end{equation}
where $a'$ represents the prediction and the $f_{M}$ represents the decoding process of model $M$. 

\subsection{LATIN-Tuning}
\label{sec:latin-tuning}

\begin{figure}[t]
\centering
\small
\includegraphics[width=1\linewidth]{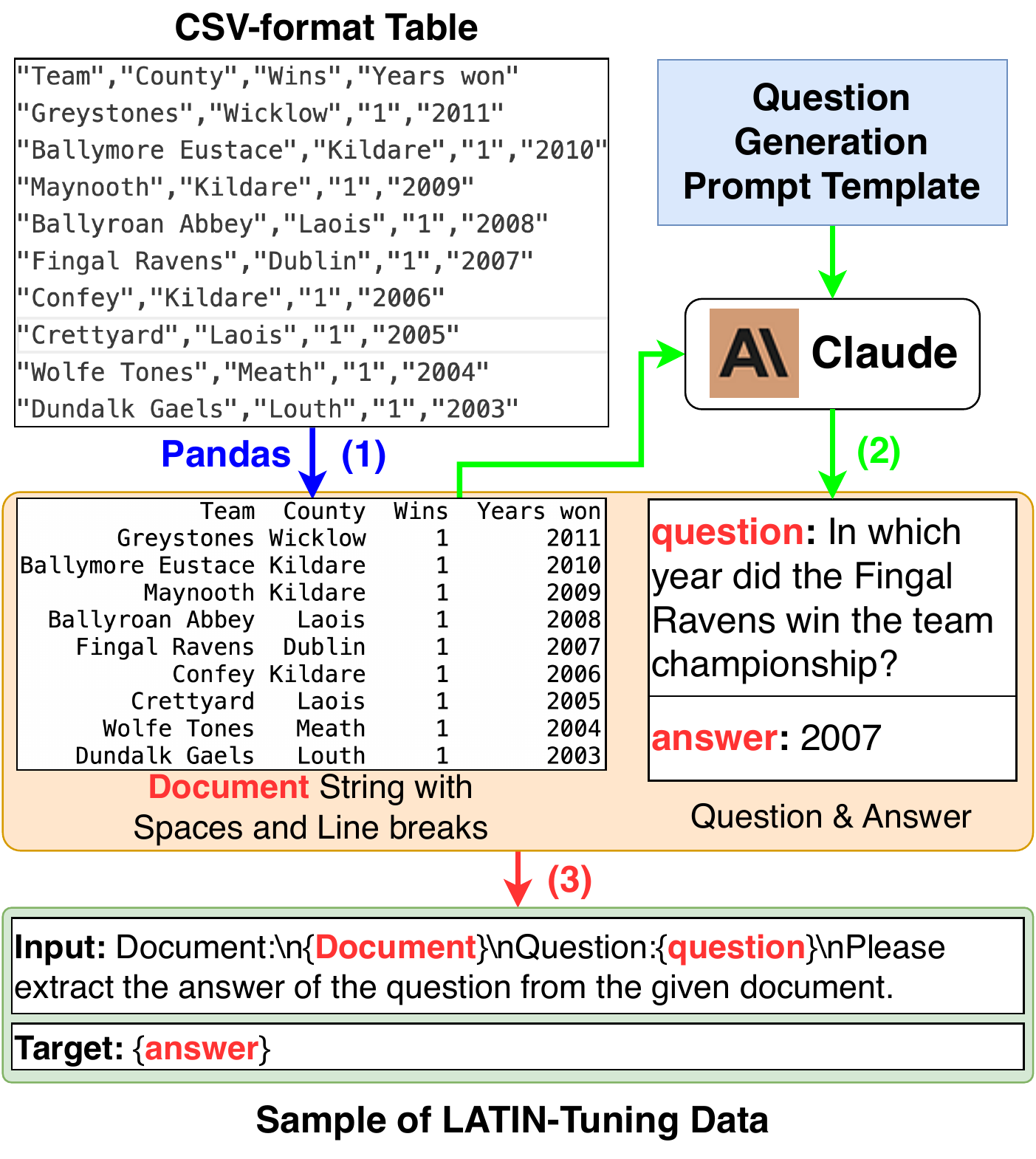}
\caption{
Construction of LATIN-Tuning data (\cref{sec:latin-tuning}).
\textcolor{blue}{
(1) Convert the CSV-format table into document string with spaces and line breaks by Pandas.
}
\textcolor{green}{
(2) Insert document string into the Question Generation Prompt Template and generate a question-answer pair by Claude.
}
\textcolor{red}{
(3) Insert document string and question into the Instruction Prompt Template to form the input, with the answer serving as the target.
}
}
\label{fig:latin-tuning}
\end{figure}

\begin{table*}[t]
\small
\centering
\begin{tabular}{@{}clccccccc@{}}
\toprule
Paradigm                      & Method                                   & Parameters               & Text         & Vision       & Layout       & Fine-tuning Set & ANLS   & $\Delta$ANLS \\ \midrule
\multirow{12}{*}{Fine-tuning} & BERT$_\text{LARGE}$                    & 340M                     &$\checkmark$ &              &              & train           & 0.6768 &              \\
                              & RoBERTa$_\text{LARGE}$                   & 355M                     & $\checkmark$ &              &              & train           & 0.6952 &              \\
                              & UniLMv2$_\text{LARGE}$                   & 340M                     & $\checkmark$ &              &              & train           & 0.7709 &              \\ \cmidrule(l){2-9} 
                              & LayoutLM$_\text{LARGE}$                 & 343M                     & $\checkmark$ &              & $\checkmark$ & train           & 0.7259 &              \\
                              & LayoutLMv2$_\text{LARGE}$      & 426M                     & $\checkmark$ & $\checkmark$ & $\checkmark$ & train           & 0.8348 &              \\
                              & LayoutLMv3$_\text{LARGE}$                & 368M                     & $\checkmark$ & $\checkmark$ & $\checkmark$ & train           & 0.8337 &              \\
                              & ERNIE-Layout$_\text{LARGE}$              & 507M                     & $\checkmark$ & $\checkmark$ & $\checkmark$ & train           & 0.8321 &              \\
                              & LayoutLMv2$_\text{LARGE}$          & 426M                     & $\checkmark$ & $\checkmark$ & $\checkmark$ & train + dev     & 0.8529 &              \\
                              & ERNIE-Layout$_\text{LARGE}$              & 507M                     & $\checkmark$ & $\checkmark$ & $\checkmark$ & train + dev     & 0.8486 &              \\
\midrule
\multirow{7}{*}{Zero-shot}    & Alpaca+Plain Prompt                    & \multirow{2}{*}{7B}      & $\checkmark$ &              &              & -               & 0.3567 &              \\
                              & Alpaca+LATIN-Prompt                    &                          & $\checkmark$ &              & $\checkmark$ & -               & 0.4200 & +0.0633      \\ \cmidrule(l){2-9} 
                              & Claude+Plain Prompt                    & \multirow{2}{*}{Unknown} & $\checkmark$ &              &              & -               & 0.2298 &              \\
                              & Claude+LATIN-Prompt                    &                          & $\checkmark$ &              & $\checkmark$ & -               & 0.8336 & +0.6038      \\ \cmidrule(l){2-9} 
                              & ChatGPT-3.5+Plain Prompt                   & \multirow{2}{*}{Unknown} & $\checkmark$ &              & $\checkmark$ & -               & 0.6866 &              \\
                              & ChatGPT-3.5+LATIN-Prompt                   &                          & $\checkmark$ &              & $\checkmark$ & -               & 0.8255 & +0.1389      \\ \cmidrule(l){2-9} 
                              & GPT-4$^*$                              & Unknown                  & \multicolumn{3}{c}{not clearly described}  & -               & 0.8840 &              \\ \bottomrule
\multicolumn{9}{l}{\parbox[c]{16cm}{\small* represents that we report the result of GPT-4 presented in OpenAI blog~\cite{OpenAIGPT4Blog2023}.
Although lacking a technical detail description, compared with Claude and ChatGPT-3.5, GPT-4 utilizes visual information.
The LATIN-Prompt is orthogonal to GPT-4 and can be used to further improve the performance of GPT-4. However, due to API permission limitations, we are unable to evaluate the performance of GPT-4 + LATIN-Prompt and leave it in future.}}
\end{tabular}
\caption{
Performance on test dataset of DocVQA.
Text, Vision, and Layout represent the modal information used by the model.
The $\Delta$ANLS represents the gain of LATIN-Prompt compared to Plain Prompt.
Unknown indicates missing relevant details.
}
\label{tab:exp_docvqa}
\end{table*}

\begin{table*}[t]
\small
\centering
\begin{tabular}{@{}cl|cc|cccc@{}}
\toprule
\multirow{2}{*}{Paradigm}    & \multirow{2}{*}{Method} & \multicolumn{2}{c|}{Overall}   & \multicolumn{4}{c}{Answer type}                                       \\
                             &                         & ANLS            & $\Delta$ANLS & Image span      & Question span   & Multiple spans  & Non span        \\ \midrule
\multirow{6}{*}{Fine-tuning} & BERT                    & 0.2078          &              & 0.2625          & 0.2333          & 0.0739          & 0.0259          \\
                             & LayoutLM          & 0.2720          &              & 0.3278          & 0.2386          & 0.0450          & 0.1371          \\
                             & LayoutLMv2              & 0.2829          &              & 0.3430          & 0.2763          & 0.0641          & 0.1114          \\
                             & BROS                    & 0.3219          &              & 0.3997          & 0.2317          & 0.1064          & 0.1068          \\
                             & pix2struct              & 0.4001          &              & 0.4308          & 0.4839          & 0.2059          & 0.3173          \\
                             & TILT                    & \textbf{0.6120} &     & \textbf{0.6765} & \textbf{0.6419} & \textbf{0.4391} & \textbf{0.3832} \\ \midrule
\multirow{4}{*}{Zero-shot}   & Claude + Plain Prompt    & 0.0798          &              & 0.0951          & 0.0913          & 0.0203          & 0.0280          \\
                             & Claude + LATIN-Prompt   & \underline{0.5451}    & +0.4653      & \underline{0.5992}    & \underline{0.5861}    &\underline{0.3985}   &\underline{0.3544}    \\ \cmidrule(l){2-8} 
                             & ChatGPT-3.5 + Plain Prompt   & 0.3335          &              & 0.3749          & 0.4505          & 0.0950          & 0.1822          \\
                             & ChatGPT-3.5 + LATIN-Prompt  & 0.4898          & +0.1563      & 0.5457          & 0.5639          & 0.3458          & 0.2798          \\ \bottomrule
\end{tabular}

\begin{tabular}{@{}l|ccccc|ccc@{}}
\toprule
\multirow{2}{*}{Method} & \multicolumn{5}{c|}{Evidence}                                                           & \multicolumn{3}{c}{Operation}                       \\
& Table/List      & Textual         & Visual object   & Figure          & Map             & Comparison      & Arithmetic      & Counting        \\ \midrule
BERT                    & 0.1852          & 0.2995          & 0.0896          & 0.1942          & 0.1709          & 0.1805          & 0.0160          & 0.0436          \\
LayoutLM          & 0.2400          & 0.3626          & 0.1705          & 0.2551          & 0.2205          & 0.1836          & 0.1559          & 0.1140          \\
LayoutLMv2              & 0.2449          & 0.3855          & 0.1440          & 0.2601          & 0.3110          & 0.1897          & 0.1130          & 0.1158          \\
BROS                    & 0.2653          & 0.4488          & 0.1878          & 0.3095          & 0.3231          & 0.2020          & 0.1480          & 0.0695          \\
pix2struct              & 0.3833          & 0.5256          & 0.2572          & 0.3726          & 0.3283          & 0.2762          & 0.4198          & 0.2017          \\
TILT& \textbf{0.5917} & \textbf{0.7916} & \underline{0.4545}    & \textbf{0.5654} & \underline{0.4480}    & \textbf{0.4801} & \textbf{0.4958} & 0.2652          \\ \midrule
Claude + Plain Prompt   & 0.0849          & 0.1099          & 0.0858          & 0.0695          & 0.0496          & 0.0589          & 0.0271          & 0.0368          \\
Claude + LATIN-Prompt   & \underline{0.5421}    & \underline{0.6725}    & \textbf{0.4897} & \underline{0.5027}    & \textbf{0.4982} & \underline{0.4598}    & \underline{0.4311}    & \textbf{0.2708} \\ \cmidrule(l){1-9} 
ChatGPT-3.5 + Plain Prompt  & 0.3481          & 0.3893          & 0.3670          & 0.3114          & 0.1843          & 0.2349          & 0.1466          & 0.2320          \\
ChatGPT-3.5 + LATIN-Prompt  & 0.4917          & 0.6016          & 0.4491          & 0.4585          & 0.3614          & 0.4312          & 0.3157          & \underline{0.2660}    \\ \bottomrule
\end{tabular}
\caption{
Performance on test dataset of InfographicVQA.
The questions in InfographicVQA can be grouped according to answer type, evidence source, and operation.
We list both the overall performance of the model and its performance on different groups.
All performances are evaluated by ANLS.
The $\Delta$ANLS represents the gain of LATIN-Prompt compared to Plain Prompt.
The highest and second-highest scores are bolded and underlined.
}
\label{tab:exp_infographicvqa}
\end{table*}
\begin{table}[t]
\small
\centering
\begin{tabular}{@{}clcc@{}}
\toprule
Paradigm                      & Method                      & Setup     & ANLS            \\ \midrule
\multirow{11}{*}{Fine-tuning} & \multirow{2}{*}{BERT}       & Max Conf  & 0.5347          \\
                              &                             & Concat    & 0.4183          \\ \cmidrule(l){2-4} 
                              & \multirow{2}{*}{Longformer} & Max Conf  & 0.5506          \\
                              &                             & Concat    & 0.5287          \\ \cmidrule(l){2-4} 
                              & \multirow{2}{*}{Big Bird}   & Max Conf  & 0.5854          \\
                              &                             & Concat    & 0.4929          \\ \cmidrule(l){2-4} 
                              & \multirow{2}{*}{LayoutLMv3} & Max Conf  & 0.5513          \\
                              &                             & Concat    & 0.4538          \\ \cmidrule(l){2-4} 
                              & \multirow{2}{*}{T5}         & Max Conf  & 0.4028          \\
                              &                             & Concat    & 0.5050          \\ \cmidrule(l){2-4} 
                              & Hi-VT5                      & Multipage & \textbf{0.6201} \\ \midrule
Zero-shot                     & Claude+LATIN-Prompt      & Max Conf  & \underline{0.6129}    \\ \bottomrule
\end{tabular}
\caption{
Performance on test dataset of MP-DocVQA.
Concat indicates concatenating multi-page content.
MaxConf indicates processing pages separately and choosing an answer based on confidence.
We only evaluate Claude with LATIN-Prompt in MaxConf setting because Plain Prompt does not apply to MaxConf, and Concat exceeds the length limit.
}
\label{tab:exp_mpdocvqa}
\end{table}
Although instruction-tuning models like Claude and ChatGPT can comprehend and utilize LATIN-Prompt well, we found that the performance of smaller models like Alpaca (7B) was not up to par.
So we propose LATIN-Tuning to enhance their ability to comprehend layout by spaces and line breaks.
As shown in \cref{fig:latin-tuning}, we employ the Pandas\footnote{https://pandas.pydata.org} and Claude to construct instruction-tuning dataset from CSV-format tables.
The process is as follows:

(1) For each CSV table, we convert it into document string with spaces and line breaks by Pandas.
Please refer to the appendix for the code implementation.
(2) We insert the document string into the Question Generation Prompt Template and generate a question-answer pair by Claude.
(3) We insert document string and question into the Instruction Prompt Template to form the input, with the answer serving as the target.
Refer to the appendix for the details of Question Generation Prompt Template and Instruction Prompt Template.

At last, we fine-tune the Alpaca on the instruction-tuning dataset to enhance its ability to comprehend layout by spaces and line breaks.

\section{Experiment}
\label{sec:exp}
\begin{figure}[t]
\centering
\small
\includegraphics[width=1\linewidth]{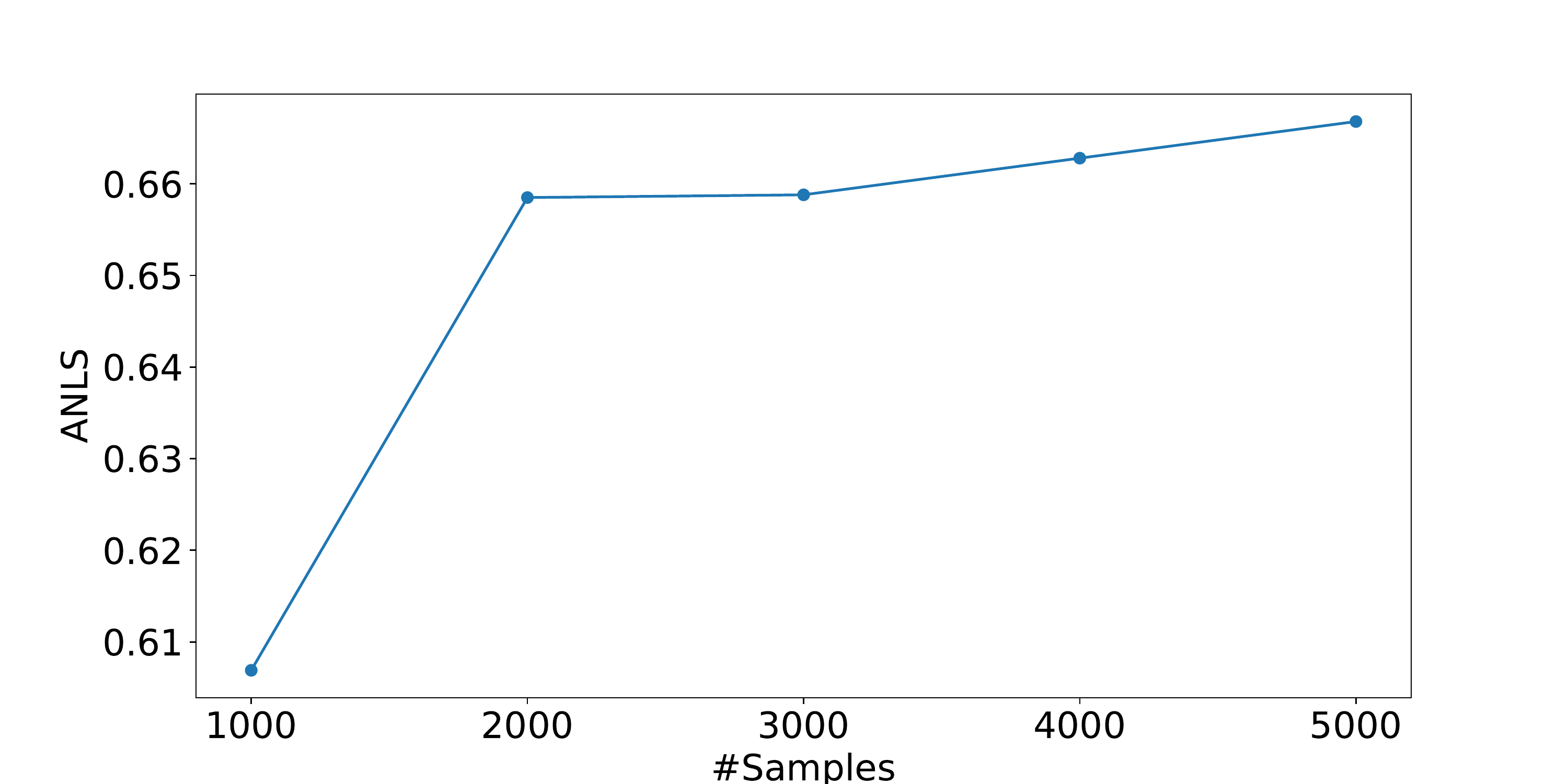}
\caption{
The impact of the size of instruction fine-tuning dataset on LATIN-Tuning.
The performance of LATIN-Tuning improves as the number of samples increases.
}
\label{fig:exp_ablation_latin_tuning}
\end{figure}

\begin{figure*}[t]
\small
\centering
\includegraphics[width=1\linewidth]{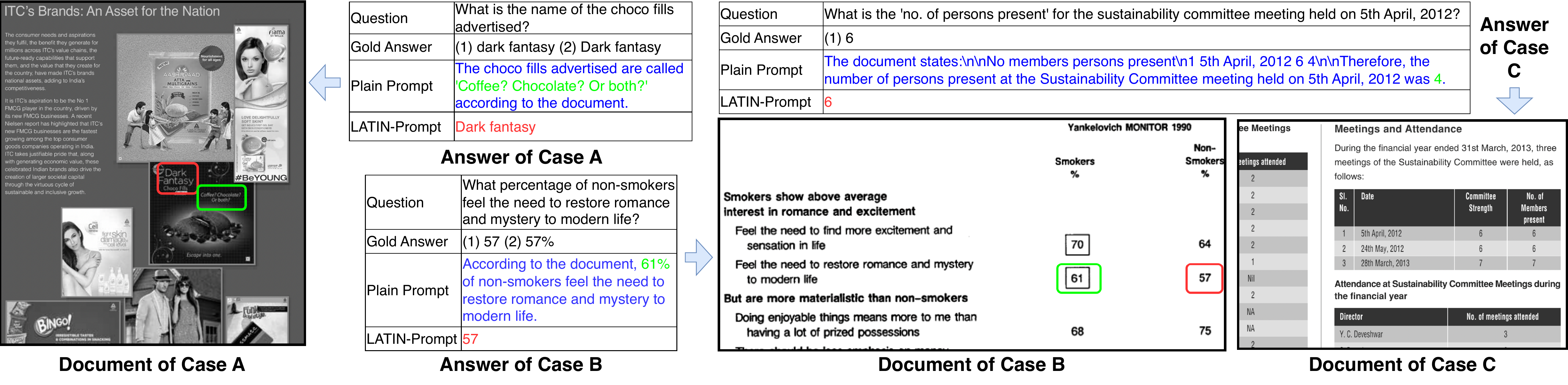}
\caption{
Case study of Claude on DocVQA.
Due to the lack of task instruction, Plain Prompt generates unnecessary words (\textcolor{blue}{in blue}), violating the extraction requirement.
Moreover, Plain Prompt cannot capture layout information and generates incorrect answers (\textcolor{green}{in green}).
In Case (A), it regards "Coffee" and "Chocolate", which are semantically similar to "choco", as the answer.
In Case (B), it regards "61" as the answer since it directly follows "romance and mystery".
In Case (C), it fails to comprehend the document and engages in erroneous reasoning.
In contrast, LATIN-Prompt can understand layout relationships and generate the correct answer (\textcolor{red}{in red}).
The document images of these cases are complex.
Due to limited space, we only display a portion of the original document images here.
Please refer to the appendix for the original document images and more cases.
}
\label{fig:case}
\end{figure*}

\subsection{Experiment Settings}
\label{sec:exp_settings}
\paragraph{Datasets}
We evaluate our method on three document image question answering datasets:
DocVQA~\cite{mathewDocVQA2021} is an extractive question answering task and consists of 50,000 questions defined on 12,767 document images;
InfographicVQA~\cite{mathewInfographicVQA2021} consists of 5,485 infographics, which convey information through text, graphics, and visual elements together;
MP-DocVQA~\cite{titoMulti-PageDocVQAHi-VT52023} extends DocVQA to more realistic multi-page scenarios where a document typically consists of multiple pages that should be processed together.
Following common practice, we use Azure OCR results for DocVQA provided by DUE~\cite{borchmannDUEEndtoEndDocument2021} and use offical OCR results\footnote{https://rrc.cvc.uab.es/?ch=17\&com=introduction} for InfographicVQA and MP-DocVQA.
For all datasets, we adopt the Average Normalized Levenshtein Similarity (ANLS)~\cite{bitenST-VQA2019} as the evaluation metric.

\paragraph{LATIN-Prompt Baselines}
To evaluate zero-shot performance of LATIN-Prompt, we compare it with Plain Prompt on three instruction-tuning language models: Claude\footnote{We use the claude-v1.3 API.}, ChatGPT-3.5\footnote{We use the gpt-3.5-turbo API from Azure OpenAI.}, and Alpaca.
The template of Plain Prompt is as follows: ``Document: \{document\}$\backslash$nQuestion: \{question\}$\backslash$nDirectly extract the answer of the question from the document.$\backslash$nAnswer:'', where \{document\} and \{question\} are placeholders of original text segments from OCR tools and question.
We also compare LATIN-Prompt's zero-shot performance with fine-tuning performance of pre-training-fine-tuning methods.
Moreover, we report the result of multimodal GPT-4 presented in OpenAI blog~\cite{OpenAIGPT4Blog2023}.
Due to API permission restrictions, we leave the exploration of GPT-4+LATIN-Prompt as future work.

We only evaluate Alpaca on DocVQA because the other two tasks are too complex for it.
Alpaca cannot follow the task instruction of these two tasks.
In fact, Alpaca performs poorly on InfographicVQA (refer to \cref{tab:exp_latin-tuning}) and cannot generate answers meeting the format requirement of MP-DocVQA.
We exclude the ChatGPT-3.5 on MP-DocVQA because it needs to process too many document pages and the experimental cost exceeds the range we can afford.

\paragraph{LATIN-Tuning} We randomly sample 5000 CSV-format tables from the WikiTableQuestions~\cite{pasupatWikiTableQuestions2015} with replacement to create the instruction-tuning dataset.
We fine-tune the Alpaca on the created dataset for 3 epochs using the AdamW~\cite{loshchilovAdamW2018} optimizer with a warmup ratio of 0.03 following the Alpaca~\cite{StanfordAlpacaInstructionfollowing2023}.
We use a batch size of $64$ and a learning rate of $2e-5$.
The resulting model is denoted as Alpaca+LATIN-Tuning and we compare it with Alpaca to evaluate the performance of LATIN-Tuning.

\subsection{Performance of LATIN-Prompt}
\Cref{tab:exp_docvqa} presents the experimental results on DocVQA.
(1) In the pre-training-fine-tuning paradigm, the layout-aware multimodal pre-trained model performs better than the pure language model.
(2) Further, increasing the amount of fine-tuning data can improve the performance of models.
(3) The instruction-tuning language models perform poorly with the plain prompt based on the original text segments obtained by OCR tools.
(4) The LATIN-Prompt proposed in this paper significantly improves the zero-shot performance of instruction-tuning language models.
It enables the zero-shot performance of Claude and ChatGPT-3.5 to significantly outperform the fine-tuned layout-aware LayoutLM.
In addition, despite only using text and layout information, their zero-shot performance is comparable to the performance of fine-tuned layout-aware multimodal pre-trained models.
(5) Although unknown, the number of parameters of Claude and GPT-3.5 should be much larger than that of Alpaca.
The experimental results show that the final zero-shot performance is positively correlated with the size and ability of the instruction-tuning models.
(6) The zero-shot performance of GPT-4 matched the best fine-tuned performance.
Although lacking a technical detail description, compared with Claude and GPT-3.5, GPT-4 utilizes visual information, reflecting the importance of visual information for document image understanding.
LATIN-Prompt is orthogonal to GPT-4.
However, due to API permission restrictions, we can only leave arming GPT-4 with the LATIN-Prompt to the future.

\Cref{tab:exp_infographicvqa} presents results on InfographicVQA.
Experimental results show that LATIN-Prompt enable the zero-shot performance of Claude and GPT-3.5 to exceed the performance of all fine-tuned baselines except TILT.
We find that Claude performs poorly with Plain Prompt, but its performance improves significantly when using LATIN-Prompt.

\Cref{tab:exp_mpdocvqa} presents results on MP-DocVQA.
It shows that, with LATIN-Prompt, Claude's zero-shot performance exceeds fine-tuning performance of Longformer~\cite{beltagyLongformer2020} and Big Bird~\cite{zaheerBigBirdTransformers2021} designed for long sequences.
Furthermore, its zero-shot performance is comparable to the fine-tuning performance of Hi-VT5~\cite{titoMulti-PageDocVQAHi-VT52023}, a layout-aware multimodal model for multi-page document images.

\subsection{Performance of LATIN-Tuning}
\begin{table}[t]
\small
\centering
\begin{tabular}{@{}lcccc@{}}
\toprule
\multicolumn{1}{c}{\multirow{2}{*}{Method}} & \multicolumn{2}{c}{DocVQA} & \multicolumn{2}{c}{InfographicVQA} \\
\multicolumn{1}{c}{}&Valid&Test&Valid&Test\\
\midrule
Alpaca&0.3506&0.3567&0.1083&0.1419\\
Alpaca+LATIN-Tuning&0.6668&0.6697&0.2873&0.3028\\
Claude&0.8311&0.8336&0.5218&0.5451\\
\bottomrule
\end{tabular}
\caption{
Effect of LATIN-Tuning. All methods are equipped with LATIN-Prompt.
}
\label{tab:exp_latin-tuning}
\end{table}
\Cref{tab:exp_latin-tuning} demonstrates that LATIN-Tuning improves the performance of Alpaca on DocVQA by $87.7\%$ and on InfographicVQA by $113\%$.
Nevertheless, its performance still lags behind Claude.
We will explore more effective instruction-tuning method in the future.

\subsection{Quantitative and Qualitative Analyses}
\begin{table}[t]
\small
\centering
\begin{tabular}{@{}lcccc@{}}
\toprule
\multicolumn{1}{c}{\multirow{2}{*}{Prompt}} & \multicolumn{2}{c}{DocVQA} & \multicolumn{2}{c}{InfographicVQA} \\
\multicolumn{1}{c}{}&Claude&ChatGPT&Claude&ChatGPT\\
\midrule
LATIN-Prompt&0.8311&0.8135&0.5218&0.4708\\
w/o Layout&0.7825&0.7491&0.4638&0.4341\\
w/o Task&0.3637&0.7561&0.1234&0.4296\\
w/o Task+Layout&0.2144&0.6795&0.0702&0.3103\\
\bottomrule
\end{tabular}
\caption{
Ablation of LATIN-Prompt on validation data of DocVQA and InfographicVQA.
}
\label{tab:ablation_study_claude}
\end{table}

\paragraph{Effect of components of LATIN-Prompt}
LATIN-Prompt consists of layout-aware document content (Layout) and task instruction (Task).
\Cref{tab:ablation_study_claude} presents the results of ablation study of LATIN-Prompt with Claude and ChatGPT-3.5 on DocVQA and InfographicVQA.
The results show that both the layout-aware document content and the task instruction can significantly improve the zero-shot performance of Claude and ChatGPT-3.5.
The improvement brought by task instruction is more significant in Claude because it ensures that the format of the answers generated by the model meets the task requirements.
On the basis of the correct format, the layout-aware document content further improves the performance of the model because it enables the model to utilize the layout information among text segments.

\paragraph{Effect of instruction-tuning data size for LATIN-Tuning}
\Cref{fig:exp_ablation_latin_tuning} shows that the performance of LATIN-Tuning improves as the number of samples increases.
The improvement rate slows down when the sample count exceeds 2000.

\paragraph{Case study of LATIN-Prompt}
\Cref{fig:case} provides cases of Claude on DocVQA.
Compared with Plain Prompt, LATIN-Prompt enables the model to comprehend layout more effectively and generate answers meeting the format requirement.

\paragraph{Case study of LATIN-Tuning}
Case study on DocVQA shows that LATIN-Tuning enables Alpaca to understand layout by spaces.
Please refer to the appendix for details.


\section{Conclusion}
In this work, we point a new perspective for comprehending layout information within document images.
Instead of capturing layout by coordinate of bounding boxes, we find that instruction-tuning language models like Claude and ChatGPT can understand layout by spaces and line breaks.
Based on this observation, we propose LATIN-Prompt and it enables zero-shot performance of Claude and ChatGPT to be comparable to the fine-tuning performance of SOTAs on document image question answering.
Moreover, we propose LATIN-Tuning, which enhances the ability of Alpaca to comprehend layout by spaces and line breaks.
In the future, we will explore to incorporate visual information into LATIN-Prompt and create more effective instruction-tuning dataset for LATIN-Tuning.

\bibliography{aaai24}

\appendix

\section{Prompt Templates}
\begin{table*}[ht]
    \small
    \centering
    \begin{tabular}{cl}
    \toprule
    \#Line& Prompt\\
    \midrule
    1& You are asked to answer questions asked on a document image.\\
    2& \parbox[c]{16cm}{
    The answers to questions are short text spans taken verbatim from the document.
    This means that the answers comprise a set of contiguous text tokens present in the document.}\\
    3& Document:\\
    4& \textcolor{red}{\{Layout Aware Document placeholder\}}\\
    5& \\
    6& Question: \textcolor{red}{\{Question placeholder\}}\\
    7&\\
    8&Directly extract the answer of the question from the document with as few words as possible.\\
    9&\\
    10&Answer:\\
    \bottomrule
\end{tabular}
\caption{DocVQA Prompt Template. \textcolor{red}{The \{\} represents the placeholder.}}
\label{tab:docvqa_prompt}
\end{table*}

\begin{table*}[ht]
\small
\centering
\begin{tabular}{cl}
\toprule
\#Line& Prompt\\
\midrule
1& You are asked to answer questions asked on a document image.\\
2& The answer for a question in this can be any of the following types:\\
3& 1. Answer is a piece contiguous text from the document.\\
4& \parbox[c]{16cm}{2. Answer is a list of "items" , where each item is a piece of text from the document (multiple spans).
In such cases your model/method is expected to output an answer where each item is separated by a comma and a space.
For example if the question is "What are the three common symptoms of COVID-19?" Answer must be in the format "fever, dry cough, tiredness".
In such cases "and" should not be used to connect last item and the penultimate item and a space after the comma is required so that your answer match exactly with the ground truth.}\\
5& 3. Answer is a contiguous piece of text from the question itself (a span from the question)\\
6& \parbox[c]{16cm}{4. Answer is a number ( for example "2", "2.5", "2\%", " 2/3" etc..).
For example there are questions asking for count of something or cases where answer is sum of two values given in the image.}\\
7& Document:\\
8& \textcolor{red}{\{Layout Aware Document placeholder\}}\\
9& \\
10& Question: \textcolor{red}{\{Question placeholder\}}\\
11& \\
12& Directly answer the answer of the question from the document with as few words as possible.\\
13& \\
14& Answer:\\
\bottomrule
\end{tabular}
\caption{
InfographicVQA Prompt Template. \textcolor{red}{The \{\} represents the placeholder.}
}
\label{tab:infographicvqa_prompt}
\end{table*}

\begin{table*}[ht]
\small
\centering
\begin{tabular}{cl}
\toprule
\#Line& Prompt\\
\midrule
1& You are asked to answer questions asked on a document image.\\
2& \parbox[c]{16cm}{
The answers to questions are short text spans taken verbatim from the document.
This means that the answers comprise a set of contiguous text tokens present in the document.}\\
3& Document:\\
4& \textcolor{red}{\{Layout Aware Document placeholder\}}\\
5& \\
6& Question: \textcolor{red}{\{Question placeholder\}}\\
7& \\
8& Directly extract the answer of the question from the document with as few words as possible.\\
9& \\
10& You also need to output your confidence in the answer, which must be an integer between 0-100. \\
11& \parbox[c]{16cm}{The output format is as follows, where [] indicates a placeholder and does not need to be actually output:}
\\
12& [Confidence score], [Extracted Answer] \\
\bottomrule
\end{tabular}
\caption{MP-DocVQA Prompt Template. \textcolor{red}{The \{\} represents the placeholder.}}
\label{tab:mpdocvqa_prompt}
\end{table*}

\begin{table*}[ht]
\small
\centering
\begin{tabular}{cl}
\toprule
\#Line& Prompt\\
\midrule
1& Document:\\
2& \textcolor{red}{\{Document string with spaces and line from CSV table\}}\\
3& \parbox[c]{16cm}{
Randomly generate a question and corresponding answer for the above document.
The answer to the question must be unique, and must be extracted from the document.
To answer this question, the layout of the document must be understood.
The output should be in the following format without any other text:
}\\
4& Question: [Question content]\\
5& Answer: [Answer content]\\
\bottomrule
\end{tabular}
\caption{Instruction Prompt Template. \textcolor{red}{The \{\} represents the placeholder.}
}
\label{tab:question_prompt}
\end{table*}

\begin{table*}[ht]
\small
\centering
\begin{tabular}{cl}
\toprule
\#Line& Prompt\\
\midrule
1& Document:\\
2& \textcolor{red}{\{Document string with spaces and line from CSV table\}}\\
3& Question: \textcolor{red}{\{Question\}}\\
4& Please extract the answer of the question from the given document.\\
\bottomrule
\end{tabular}
\caption{Question Generation Prompt Template. \textcolor{red}{The \{\} represents the placeholder.}
}
\label{tab:instruction_prompt}
\end{table*}

\Cref{tab:docvqa_prompt} shows the prompt template for DocVQA\cite{mathewDocVQA2021}.
In the first and second lines, we explain the meaning of extraction in detail to the model according to the task description in DocVQA~\cite{mathewDocVQA2021}.
Lines 3 to 6 provide placeholders for the layout-aware document and question.
To avoid the model forgetting its task due to the interference of document content, the 8th line summarizes and reiterates the task requirements.

\Cref{tab:infographicvqa_prompt} shows the prompt template for InfographicVQA\cite{mathewInfographicVQA2021}.
Compared with DocVQA, InfographicVQA has more complex answer sources.
We describe the answer requirements in detail in lines 1 to 6, and the rest is similar to the DocVQA prompt template.

\Cref{tab:mpdocvqa_prompt} shows the prompt template for MP-DocVQA \cite{titoMulti-PageDocVQAHi-VT52023}.
MP-DocVQA is a multi-page question answering task, in which each question has multiple candidate page images, and we adopt the Max Confidence (Max Conf)~\cite{mathewDocVQA2021} setup to solve it.
Specifically, the model extracts one answer on each page and gives confidence to this answer.
We choose the answer with the highest confidence as the final prediction.
Lines 10 to 12 of \cref{tab:mpdocvqa_prompt} instruct the model to output the answer and confidence simultaneously, and the rest is similar to \cref{tab:docvqa_prompt}.

\Cref{tab:question_prompt} and \Cref{tab:instruction_prompt} provide the details of Question Generation Prompt Template and Instruction Prompt Template in LATIN-Tuning. 

\section{Datasets}
\label{sec:exp_settings}
To evaluate LATIN-Prompt, we conduct experiments on three document image question answering datasets, including DocVQA~\cite{mathewDocVQA2021}, InfographicVQA~\cite{mathewInfographicVQA2021}, and MP-DocVQA~\cite{titoMulti-PageDocVQAHi-VT52023}.

The DocVQA is an extractive question answering task and consists of 50,000 questions defined on 12,767 document images.
The train split has 39,463 questions, the validation split has 5,349 questions, and the test split has 5,188 questions.
DocVQA contains a large number of questions related to forms, layouts, tables and lists in the image, posing high requirements on the model's ability to understand document image layouts.

The InfographicVQA consists of 5,485 infographics, which convey information through text, graphics, and visual elements together.
Compared with DocVQA, InfographicVQA emphasizes questions that require basic reasoning and arithmetic skills, and has more complex answer sources, including Document(Image)-Span, Question-Span, Multi-Span and Non-extractive.

MP-DocVQA extends DocVQA to more realistic multi-page scenarios where a document typically consists of multiple pages that should be processed together.
It comprises 46,000 questions posed over 48,000 scanned pages belonging to 6,000 industry documents, and the page images contain diverse layouts.
The variability between documents in MP-DocVQA is very high.
The number of pages in each document varies from 1 to 20, and the number of recognized OCR words varies from 1 to 42,313.

Following common practice, we use Azure OCR results for the DocVQA provided by DUE~\cite{borchmannDUEEndtoEndDocument2021}.
For the InfographicVQA and MP-DocVQA, we use offical OCR results\footnote{https://rrc.cvc.uab.es/?ch=17\&com=introduction}.

\section{Case Study}
\begin{figure}[h]
\small
\centering
\includegraphics[width=1\linewidth]{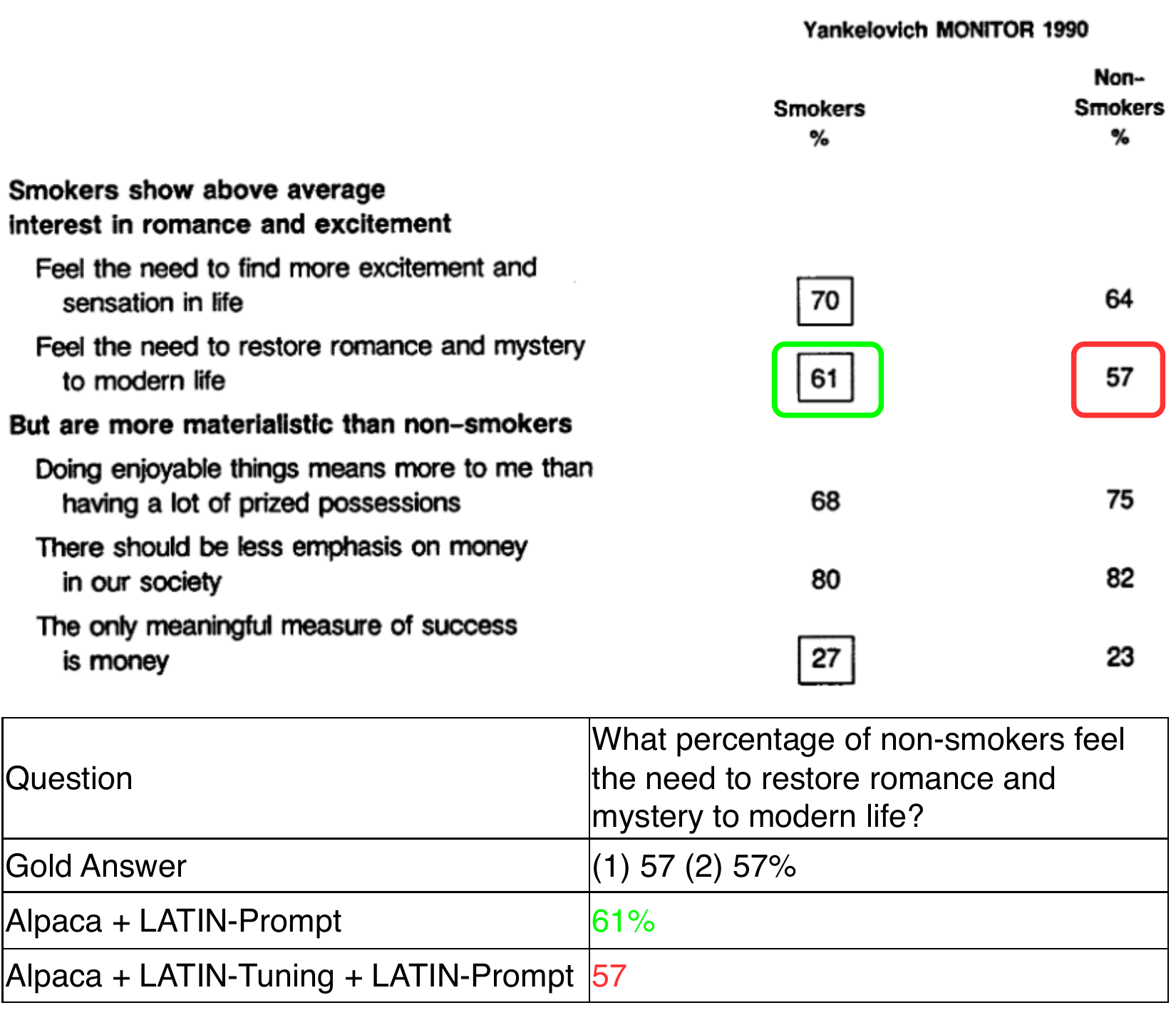}
\caption{
Case study of LATIN-Tuning.
}
\label{fig:LATIN-Tuning_case}
\end{figure}

\Cref{fig:LATIN-Tuning_case} provides a case to compare the performance of Alpaca and Alpaca + LATIN-Tuning.
After LATIN-Tuning, the Alpaca can comprehend layout more effectively.

\end{document}